\pdfoutput=1

\documentclass[11pt]{article}
\usepackage{amsmath}
\usepackage{amssymb}
\usepackage{booktabs}
\usepackage{siunitx}
\usepackage{array,multirow,graphicx}
\usepackage{comment}
\usepackage{hyperref}
\usepackage{xurl}
\usepackage{algorithm}
\usepackage{algorithmic}
\usepackage{graphicx}
\usepackage{xcolor}
\usepackage{xspace}
\usepackage{enumitem}

\usepackage{pifont}
\newcommand{\xmark}{\ding{55}}%
\usepackage{color, colortbl}

\newcommand{\blaser}{\textsc{blaser}\xspace}
\newcommand{\blaseruns}{$\textsc{blaser}_\text{u}$\xspace}
\newcommand{\blasersup}{$\textsc{blaser}_\text{s}$\xspace}

\newcommand{\chrf}{\textsc{chrf+}\xspace}
\newcommand{\bertscore}{\textsc{bertscore}\xspace}
\newcommand{\asrbleu}{\textsc{asr-bleu}\xspace}
\newcommand{\asrsentbleu}{\textsc{asr-sentbleu}\xspace}
\newcommand{\asrchrf}{\textsc{asr-chrf+}\xspace}
\newcommand{\asrbertscore}{\textsc{asr-bertscore}\xspace}
\newcommand{\asrcomet}{\textsc{asr-comet}\xspace}
\newcommand{\laser}{\textsc{laser}\xspace}
\newcommand{\comet}{\textsc{comet}\xspace}
\newcommand{\bleu}{\textsc{bleu}\xspace}
\newcommand{\sentbleu}{\textsc{sentbleu}\xspace}

\newcommand{\audio}{speech segment\xspace}
\newcommand{\audios}{speech segments\xspace}

\newcommand{\mustc}{\mbox{MusT-C}\xspace}

\usepackage{lipsum}

\newcommand\blfootnote[1]{%
  \begingroup
  \renewcommand\thefootnote{}\footnote{#1}%
  \addtocounter{footnote}{-1}%
  \endgroup
}

\usepackage{acl}

\usepackage{times}
\usepackage{latexsym}

\usepackage[T1]{fontenc}

\usepackage[utf8]{inputenc}

\usepackage{microtype}

\title{BLASER: A Text-Free Speech-to-Speech Translation Evaluation Metric}

\author{Mingda Chen, Paul-Ambroise Duquenne, Pierre Andrews, Justine Kao, \\ \bf{Alexandre Mourachko, Holger Schwenk$^*$, Marta R. Costa-juss\`a$^*$} \\
 Meta AI\\
  \texttt{\{mingdachen,padqn,mortimer,jtk,\}@meta.com}\\
  \texttt{\{alexmourachko,schwenk,costajussa\}@meta.com} \\}

\begin{document}
\maketitle
\begin{abstract}
\blfootnote{$^*$ Equal Research Leadership Contribution}

End-to-End speech-to-speech translation (S2ST) is generally evaluated with text-based metrics. 
This means that generated speech has to be automatically transcribed, making the evaluation dependent on the availability and quality of automatic speech recognition (ASR) systems. 

In this paper, we propose a text-free evaluation metric for end-to-end S2ST, named \blaser{}, to avoid the dependency on ASR systems. \blaser{} leverages a multilingual multimodal encoder to directly encode the \audios for source input, translation output and reference into a shared embedding space and computes a score of the translation quality that can be used as a proxy to human evaluation.
To evaluate our approach, we construct training and evaluation sets from more than 40k human annotations covering seven language directions. The best results of \blaser{} are achieved by training with supervision from human rating scores. We show that when evaluated at the sentence level, \blaser{} correlates significantly better with human judgment compared to ASR-dependent  metrics including \asrsentbleu in all translation directions and \asrcomet
in five of them. Our analysis shows combining speech and text as inputs to \blaser{} does not increase the correlation with human scores, but best correlations are achieved when using speech, which motivates the goal of our research. Moreover, we show that using ASR for references is detrimental for text-based metrics. \footnote{Code is available at \url{https://github.com/facebookresearch/stopes}}

\end{abstract}

\section{Introduction}

Speech-to-Speech translation seeks to translate \audios from one language into another. Historically, it has been implemented and evaluated as a concatenation of three systems: automatic speech recognition (ASR), machine translation (MT) and text-to-speech (TTS) \cite{lavie,lazzari-2006-tc}.
In recent years, there has been increasing interest in end-to-end approaches \cite{Jia2019DirectST,lee-etal-2022-direct}.
While end-to-end S2ST is becoming popular, researchers still rely on text-based metrics to evaluate model performance by automatically transcribing the generated \audios  \cite{Jia2019DirectST}.
These cascaded metrics rely on ASR systems, which for a given language may not have enough quality or may not even be available \cite{DBLP:journals/corr/abs-2111-03945}.  
They are also inappropriate for languages lacking standardized writing systems \cite{salesky2021assessing}, like Hokkien or Algerian Arabic.

In this work, we propose the text-free metric \blaser for S2ST evaluation, sidestepping the dependency on ASR systems. In particular, we use \laser encoders that support multiple languages and modalities including text \citep{heffernan2022bitext} and speech \citep{NEURIPS2021_8466f9ac}.
We use the \laser encoders to directly embed \audios into vectors and compute a score estimating the quality of generation. We then construct training and evaluation datasets from more than 40k human annotations, covering seven language directions (Spanish$\leftrightarrow$English, French$\leftrightarrow$English, Russian$\rightarrow$English, Hokkien$\rightarrow$English, and English$\rightarrow$German). We evaluate \blaser on these datasets on the popular benchmark of \mustc \cite{di-gangi-etal-2019-must}. 
We also benchmark several strong ASR-based metrics, e.g., \asrsentbleu (i.e., sentence-level \asrbleu{} \cite{Jia2019DirectST}) and \asrcomet (i.e., applying \comet \cite{rei-etal-2020-comet} on ASR outputs). There is a recent interest of supervised evaluation metrics that are trained on human quality scores \cite{rei-etal-2020-comet}. However, these human quality scores are precious and somehow limited or nonexistent, specially for low-resource languages. Therefore, we propose both an unsupervised and a supervised version of \blaser{}.
The results show that on average both unsupervised and supervised \blaser outperform their corresponding baseline metrics. In particular, \blaser outperforms \asrcomet significantly in five language directions and obtains comparable results in two other language directions. Our analysis reveals that, while \blaser{} can use both text and speech, encoding speech data give the most significant benefits. In addition, we show that replacing human-written source input and human-written reference with ASR-generated ones hurts performance of text-based metrics, which motivates the use of modality-agnostic metrics as \blaser{}.


\section{Related Work}
\label{sec:relwork}

\paragraph{S2ST Evaluation.} Early approaches for automatic S2ST evaluation use metrics consisting of three modules where each module is used to evaluate individual component in the cascaded S2ST pipeline: e.g., \bleu and Translation Edit Rate \cite{snover-etal-2006-study} for NMT, Word Error Rate for ASR, and Mel-Cepstral Distortion \cite{kominek2008synthesizer} for TTS.
Recent approaches have been primarily focused on adapting text-based metrics for end-to-end S2ST
\cite{Jia2019DirectST,lee-etal-2022-direct}.
In contrast to these works, we propose a text-free metric.


\paragraph{MT Metrics.} There is a huge amount of literature in automatic machine translation evaluation in the area of natural language processing \cite[\emph{inter alia}]{papineni-etal-2002-bleu,denkowski-lavie-2014-meteor,popovic-2015-chrf}.
Recent methods have approached this goal by using human ratings for training model-based metrics, such as \comet{}, \bertscore{} \cite{Zhang*2020BERTScore:} and \textsc{bleurt} \cite{sellam-etal-2020-bleurt}. These metrics have achieved remarkable performance on text
\cite{freitag-etal-2021-results,kocmi-etal-2021-ship}.

\paragraph{Speech Metrics.} 

Our work involves computing semantic similarity of \audios to evaluate translation quality. It is thus related to reference-based automatic evaluation metrics for TTS where the metrics seek to measure the quality of generated \audios given reference \audios e.g., Mel-Cepstral Distortion, Gross
Pitch Error \cite{NAKATANI2008203} and other model-based metrics \cite{Bińkowski2020High}. Unlike our work, these metrics primarily focus on the \emph{naturalness} of synthesized speech.

Contemporaneous to this work, \citet{besacier2022textless} propose a text-free metric for comparing two \audios in the same language. Their work limits to comparing English speech data and they do not cover multilingual S2ST evaluation. Their work is based on synthetic datasets where ratings are generated by automatic text-based measures as opposed to human annotators. Differently, we cover S2ST evaluation and we show how our metric correlates with human annotations and how it improves over text-based metrics. 

\paragraph{Speech and/or Text Representations.} There is a large body of research on learning multilingual text embeddings for various downstream tasks. LabSE \cite{feng-etal-2022-language}, SentenceBERT \cite{reimers-2019-sentence-bert}, mUSE \cite{yang-etal-2020-multilingual} and LASER \cite{artetxe-schwenk-2019-massively,heffernan2022bitext} are popular encoders that capture the semantic information of a sentence into fixed size vector representations. In the speech modality, approaches such as wav2vec 2.0 \cite{baevski2020wav2vec} or Hubert \cite{hsu2021hubert} allow learning embeddings at acoustic-frame level. 

There has recently been increased interest in aligned speech-text representations such as mSLAM \cite{bapna2022mslam}, MAESTRO \cite{chen2022maestro}, SAMU-XLSR \cite{khurana2022samu}, and \laser{} \cite{duquenne2022speechmatrix}.
While our approach could accommodate any speech representation architecture given the right pooling strategy, we chose \laser{} in this work for three reasons. \textbf{(1)} The encoders modules are freely-available; 
(2) the \laser{} embedding space can easily be extended to new languages at a minimal cost: contrary to most multilingual encoders, the teacher-student approach does not require the whole embedding space to be retrained after including data for the new language. This makes \blaser{} virtually usable for any language in the future  \textbf{(3)} the embedding space could potentially be extended to any new modality meaningful to translation use cases. 
\section{Approach} 

The underlying idea of our approach is to leverage the similarity between \audios without requiring intermediate textual representations. 
Compared to ASR-based metrics, the advantage of \blaser{} is that it is text-free. In particular, given the source input speech, the translated output speech of a S2ST model, and the reference speech segment, respectively, we embed them into vectors $h_\text{src}$, $h_\text{mt}$, and $h_\text{ref}$.
These embeddings are combined and \blaser{} predicts a score for each translation output,
where higher 
scores suggest better translation quality.\footnote{A straightforward corpus-level score could be obtained via averaging over sentence-level scores, which can be used to compare different S2ST models, similar to metrics like \bleu.}
%

The effectiveness of \blaser depends on the quality of vector representations encoded from \audios: it requires rich semantic information to be encoded in the speech embeddings. In this work, we use \laser speech encoders \cite{duquenne2022speechmatrix}, which we describe below. We note that our approach is generic and can be extended to other encoders.


We study \blaser under the unsupervised and the supervised settings, which allows it to exploit the information of human ratings, if available.

\subsection{Background: \laser Encoders}

The LASER encoder was initially trained in a sequence-to-sequence model \cite{schwenk-douze-2017-learning} and supported 93 languages in its follow-up publications \cite{artetxe-schwenk-2019-massively}. In recent work, a teacher-student approach was applied to incorporate more languages \cite{heffernan2022bitext} and to extend the model to the speech modality \cite{NEURIPS2021_8466f9ac}. All these encoders use the same teacher model and are mutually compatible. The embeddings are of dimension 1024. The reader is referred to these papers for a detailed description. These LASER encoders were successfully applied to automatically mine semantically similar sentences, in the text \cite{costa2022no} and speech domain \cite{duquenne2022speechmatrix}.




\subsection{Unsupervised \blaser}

In the unsupervised setting, we directly compute the cosine similarities between $h_\text{src}$ and $h_\text{mt}$, and $h_\text{ref}$ and $h_\text{mt}$.
Formally, this metric is defined as follows:
%
\begin{equation}
\blaser{}_\text{u} = \frac{\text{cos}(h_\text{src}, h_\text{mt}) +
\text{cos} (h_\text{ref}, h_\text{mt})}{2}	
\label{eq:1}
\end{equation}
\noindent where $\text{cos}(\cdot,\cdot)$ is the cosine similarity function.

\subsection{Supervised \blaser}




Previous work has shown that evaluation metrics (e.g. \cite{rei-etal-2021-references}) can take advantage of human ratings for training. We follow \comet{} \cite{rei-etal-2020-comet} and \textsc{RUSE} \cite{shimanaka-etal-2018-ruse} and use the following features:
\begin{itemize}
    \item Element-wise source product: $h_\text{src}\odot h_\text{mt}$
    \item Element-wise reference product: $h_\text{ref}\odot h_\text{mt}$
    \item Absolute element-wise source difference: $\vert h_\text{src}- h_\text{mt}\vert$
    \item Absolute element-wise reference difference: $\vert h_\text{ref}- h_\text{mt}\vert$
\end{itemize}

\begin{figure}[t]
    \centering
    \includegraphics[scale=0.6]{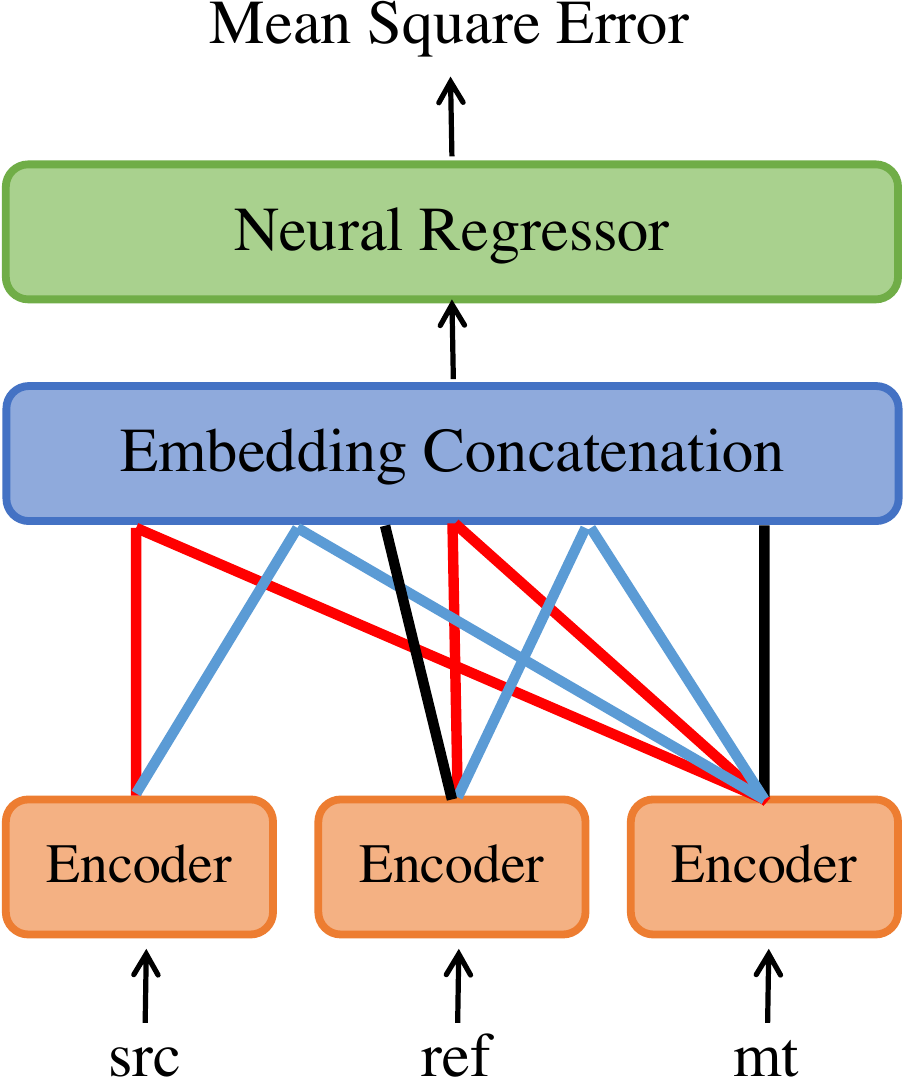}
    \caption{Diagram for supervised \blaser{}. The source input (src), reference (ref), and translation (mt) \audios are embedded into vectors using pretrained speech encoders. We create different combinations of these embeddings through element-wise product (red lines), absolute element-wise difference (blue lines), and unchanged (black lines). Combinations are concatenated as input for a neural regressor.
    We keep the encoders fixed and train the neural regressor using the Mean Square Error.
    }
    \label{fig:diagram}
\end{figure}

We concatenate these features with the embeddings of references $h_\text{ref}$ and translation outputs $h_\text{mt}$ and then use it as input for a neural regressor to predict a scalar indicating the quality of the translated speech, as shown in Figure~\ref{fig:diagram}. 
This metric corresponds to the following equation:

\begin{equation}
\begin{split}
\blaser{}_\text{s} = \text{nnet}([\text{h}_\text{ref};\text{h}_\text{mt};\text{h}_\text{src}\odot\text{h}_\text{mt};\vert\text{h}_\text{src}-\text{h}_\text{mt}\vert; \\
\text{h}_\text{ref}\odot\text{h}_\text{mt};\vert\text{h}_\text{ref}-\text{h}_\text{mt}\vert])\nonumber
\end{split}
\label{eq:2}    
\end{equation}

\noindent where $\text{nnet}(\cdot)$ is a two-layer neural network and $[\cdot;\cdot]$ represents the concatenation of vectors. We note that the dimension of concatenated input vectors to the neural regressor is 6144. The entire model except the \laser{} encoders (which are kept frozen) 
is trained by minimizing the Mean Squared Error between the \blaser{}$_\text{s}$ predicted scores and human ratings. We choose to freeze \laser encoders because (1) we do not want to break the aligned embedding space; and (2) it allows us to extend to unseen languages more easily.

\section{Experimental Framework}
\label{sec:experiments}

To show that \blaser{} is useful both in its unsupervised and supervised form, we compare it to several baseline metrics. In this section, we describe the experimental framework for doing this comparison, including the evaluation data, the training and implementation of both baseline and proposed metrics and their evaluation.


\subsection{Data} 

\begin{figure}[t]
    \centering
    \includegraphics[scale=0.6]{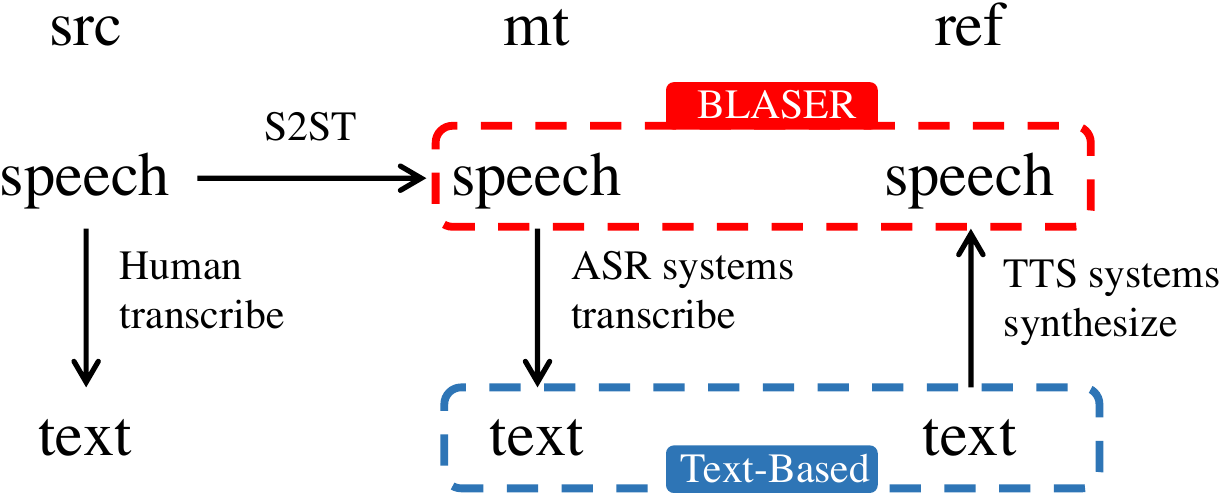}
    \caption{Diagram illustrating data sources of source input (src), reference (ref), and translation output (mt) used in this work. The source speech data and the reference text are generated/annotated by humans. We use color boxes to highlight the differences between \blaser and text-based metrics. }
    \label{fig:datasetconfig}
\end{figure}

We create training and evaluation data from \mustc \cite{di-gangi-etal-2019-must}, Multilingual TEDx \cite{salesky2021mtedx}, and TAT corpus \cite{9295019}. Given a source input from these datasets, we generate translated outputs using various S2ST models. We then conduct human evaluations to collect human ratings for generated \audios. As the datasets do not have reference \audios but provide human-written transcripts, we use TTS to synthesize speech data from these transcripts to facilitate fair comparison between our metrics and other reference-based textual metrics.
While the use of synthesized audios is disadvantageous to \blaser{},\footnote{For example, examples 2 and 3 in table \ref{tab:qual_analysis} do not correctly synthesize \textit{SMS} or \textit{PKW}.}
current benchmarks still use human-written transcripts because of the current dependence on the text-based metrics. 
We expect that, in the future, S2ST benchmarks will rely on speech references and TTS will not be needed. In this case, \blaser{} will have additional advantage over text-based metrics that will have to apply ASR to references in addition to ASR to system outputs.

Each data instance in our dataset consists of a source input, a translation output, a reference, and a human evaluation score, where the source, translation output, and reference have both speech and text. Figure~\ref{fig:datasetconfig} summarizes the data sources of these components. As follows we describe the details of each data sources.

\paragraph{Human Annotations.} We do not use crowd workers as human annotators and instead we use a vendor-managed pool of well-trained and qualified bilingual annotators who pass a qualification test for their language skills. Human annotators are instructed to rate semantic similarities between source input and generated \audios\footnote{We note that the generated \audios could be reference \audios coming from the TTS models or translated \audios coming from the S2ST models.} on a 5-point Likert scale,
where higher values are better, following annotation guidelines similar to \citet{licht-etal-2022-consistent}. More details on human evaluations are in Appendix \ref{appendix-sec:human-eval}. Each model generation has 1\textasciitilde 18 human ratings, leading to 4k\textasciitilde 20k annotations per language direction. We take medians of rating scores when there are more than one score associated with a particular model generation following \citet{costa2022no} and \citet{licht-etal-2022-consistent}.

\begin{table*}[t]
\small
    \centering
\sisetup{detect-weight,
         mode=text, 
         table-format = 5.1
         }
    \begin{tabular}{rSSSSSSS}
    \toprule
  & \textbf{es$\rightarrow$en} & \textbf{ru$\rightarrow$en} & \textbf{hk$\rightarrow$en} & \textbf{fr$\rightarrow$en} & \textbf{en$\rightarrow$de} & \textbf{en$\rightarrow$es} & \textbf{en$\rightarrow$fr} \\ \midrule
No. of annotators & 14 & 16 & 9 & 4 & 13 & 13 & 8 \\ 
No. of S2ST systems & 5 & 4 & 1 & 1 & 1 & 4 & 1 \\ 
\midrule
No. of unique source inputs & 989 & 1002 & 988 & 1015 & 2047 & 1000 & 1000 \\ 
No. of annotations & 20636 & 17908 & 6978 & 4545 & 12282 & 14817 & 4426 \\
No. of train instances & 2470 & 2004 & 0 & 0 &1023 &2000 & 0 \\ 
No. of test instances & 2475 & 2004 & 988 & 1015 &1024 &2000 &1000 \\ 
\midrule
\multicolumn{8}{l}{No. of annotations per instance}\\ 
maximum  & 6 & 6 & 18 & 6 & 6 & 6 & 6 \\ 
minimum & 1 & 1 & 4 & 1 & 6 & 1 & 2 \\ 
average & 4.2 & 4.5 & 7.1 & 4.5 & 6.0 & 3.7 & 4.4  \\ 
\bottomrule
    \end{tabular}
    \caption{Dataset Statistics. We collect human annotations for \audios generated by S2ST systems.
    }
    \label{tab:data_stat}
\end{table*}

\paragraph{Speech To Speech Translation.}
We evaluate the translation outputs generated with the following S2ST architectures:
\begin{enumerate}
    \item Cascaded two-stage models with speech-to-text translation and TTS. This system includes Spanish-English, English-French and Russian-to-English translation directions;
    \item The model presented in \citet{lee-etal-2022-textless}, which represents target speech as discrete units and uses a speech-to-unit translation model to convert source speech to target units followed by a code HiFi-GAN vocoder \cite{park2019css10,polyak21_interspeech} to convert units to waveform. This system includes English-Spanish and Russian-to-English translation directions;
    \item The model presented in \citet{inaguma2022unity}, which is similar to \citet{lee-etal-2022-textless} except that it is a two-pass direct S2ST architecture that first generates textual representations and predicts discrete acoustic units subsequently. This system includes the Spanish-to-English translation direction;
    \item The model presented in \citet{wang2022simple}, which employs mBART \cite{liu-etal-2020-multilingual-denoising} for unsupervised machine translation in their unsupervised cascaded speech-to-text translation pipeline. This system includes the Spanish-to-English translation direction.
    \item The Hokkien-to-English S2ST system is three-stage cascaded: a concatenation of Hokkien to Chinese speech-to-text translation + Chinese to English machine translation + English TTS (English text-to-unit + unit vocoder from \citet{lee-etal-2022-textless}).
    \item The English-to-German S2ST system is the MLLP-VRAIN system \cite{iranzo-sanchez-etal-2022-mllp} from IWSLT 2022 \cite{anastasopoulos-etal-2022-findings}, which is a cascaded system of separate ASR, MT, and TTS models.
\end{enumerate}

\paragraph{Automatic Speech Recognition.} For ASR, we use the open-sourced implementation in \textsc{fairseq} \cite{ott2019fairseq},\footnote{\url{https://github.com/facebookresearch/fairseq/blob/ust/examples/speech_to_speech/asr_bleu}} that provides strong models built on top of the unsupervised pretrained wav2vec \cite{schneider2019wav2vec} or XLSR \cite{conneau2020unsupervised} models. In particular, for English and Russian, we use wav2vec 2.0 large \cite{NEURIPS2020_92d1e1eb} finetuned with CTC loss \cite{10.1145/1143844.1143891}. For Hokkien, Spanish, French, and German, we use the ASR models released in \citet{chen2022speech}, \citet{grosman2021xlsr53-large-spanish}, \citet{grosman2021voxpopuli-fr-wav2vec2-large-french}, and \citet{grosman2021xlsr-1b-german}, respectively.

\paragraph{Text to Speech.} For TTS, we use the toolkit released by \citet{wang-etal-2021-fairseq}, which provides a set of recent state-of-the-art speech synthesis models.

The language directions in the final dataset are Spanish-English and French-English in both directions (i.e., en$\rightarrow$es, es$\rightarrow$en, en$\rightarrow$fr, and fr$\rightarrow$en), Russian to English (ru$\rightarrow$en), Hokkien to English (hk$\rightarrow$en) and English to German (en$\rightarrow$de). We split the data into training and test sets when there is enough data available (i.e., at least one thousand data instances for a language direction). We also make sure that there is no overlapping source inputs between train and test sets. Table~\ref{tab:data_stat} summarizes the dataset statistics.

\subsection{Baseline Metrics} We consider a variety of baseline metrics, including \bleu{} and \chrf \cite{popovic-2017-chrf}, which are standard metrics to evaluate textual similarities. While \bleu{} is by nature corpus-level, here we use the \textit{sentence}-level version due to the insufficient amount of human annotations. To differentiate these two versions, we denote the sentence-level \bleu as \sentbleu.
We also benchmark \bertscore{} \cite{Zhang*2020BERTScore:} and \comet{}, which are popular model-based metrics that correlate well with human judgments on textual data \cite{kocmi-etal-2021-ship}.\footnote{Multilingual \textsc{bleurt} \cite{pu-etal-2021-learning} reports similar performance as \comet{} on WMT metrics tasks and therefore we decided to only include \comet{} in our experiments.} We extend these metrics to speech data by using ASR systems to transcribe the machine-translated \audios. We prepend ``\textsc{asr-}'' to the beginning of the names of these metrics to indicate the use of ASR systems. Table~\ref{tab:metricsconfig} summarizes the differences among the metrics. 

Specifically, we use \bleu{}\footnote{SacreBLEU signature: nrefs:1|case:mixed|eff:yes|tok:13a|smooth:exp|version:2.2.0} and \chrf{}\footnote{SacreBLEU signature: nrefs:1|case:mixed|eff:yes|nc:6|nw:2|space:no|version:2.2.0} as implemented in SacreBLEU \cite{post-2018-call}.\footnote{\url{https://github.com/mjpost/sacrebleu}} We normalize the reference text before computing \asrsentbleu{} and \asrchrf{} to match the lowercased and punctuationless ASR output. We use the official implementations for \bertscore{}\footnote{We use language-specific configurations recommended in \url{https://github.com/Tiiiger/bert_score}} and \comet{}.\footnote{We use the ``wmt20-comet-da'' model from \url{https://github.com/Unbabel/COMET}} To form competitive baselines, we also train \comet{} from scratch on our training data ($\comet{}_\text{retrain}$) and the concatenation of our training data and the direct assessments from WMT 15-19 metrics tasks \cite{stanojevic-etal-2015-results,bojar-etal-2016-results,bojar-etal-2017-results,ma-etal-2018-results,ma-etal-2019-results} ($\comet{}_\text{retrain with WMT}$).

\begin{table}[t]
\small
    \centering
    \begin{tabular}{lcc}
    \toprule
          & req. train & req. ASR \\ 
         \multicolumn{3}{l}{Baseline Metrics}\\ \cmidrule{1-3}
        \asrsentbleu{} & \xmark  & \checkmark \\ 
        \asrchrf{} & \xmark  & \checkmark \\
        \textsc{asr-bertscore} & \checkmark & \checkmark \\
        \asrcomet{} & \checkmark & \checkmark \\
        \addlinespace[0.5em]
        \multicolumn{3}{l}{Proposed Metrics}\\\cmidrule{1-3}
        
        \blaseruns & \xmark & \xmark\\
        \blasersup & \checkmark & \xmark\\
        \bottomrule
    \end{tabular}
    \caption{Comparisons between baseline and proposed metrics regarding the dependency of training data and ASR systems. We use ``\textsc{asr-}'' to indicate that the metric depends on ASR systems to transcribe \audios.
    }
    \label{tab:metricsconfig}
\end{table}

\begin{table*}[t]
\small
    \centering
\sisetup{detect-weight,
         mode=text, 
         table-format = 1.5
         }
    \begin{tabular}{lSSSSSSSS}
    \toprule
  & \textbf{es$\rightarrow$en} & \textbf{ru$\rightarrow$en} & \textbf{hk$\rightarrow$en} & \textbf{fr$\rightarrow$en} & \textbf{en$\rightarrow$de} & \textbf{en$\rightarrow$es} & \textbf{en$\rightarrow$fr}  & \textbf{average}\\ \midrule  
 \multicolumn{9}{l}{Unsupervised Metrics} \\ \cmidrule{1-1}
\asrsentbleu{} &  0.3226 & 0.1588 & 0.2863 & 0.3277  & 0.1179 & 0.4937 & 0.4462 & 0.3076 \\ 
\asrchrf{}$^\dagger$ & 0.3910 & 0.2324 & 0.3356 & 0.3927 & 0.1469 & \bf 0.5967 & 0.5267 & 0.3746\\
\blaser{}$_\text{u}$ & \textbf{0.4970$^*$} & \textbf{0.4326$^*$} & \textbf{0.4940$^*$} & \textbf{0.4744$^*$} & \textbf{0.3148$^*$} & 0.5843 &\textbf{0.6356$^*$} &\textbf{0.4904\phantom{*}} \\ 
\addlinespace[0.5em]
\multicolumn{9}{l}{Supervised Metrics} \\ \cmidrule{1-1}
\asrbertscore{} & 0.4332 & 0.3511 & 0.4885 & 0.4184 & 0.2031 & 0.6127 & 0.6216 & 0.4469\\
\asrcomet{} & 0.5238 &0.3988 &0.5138 & 0.5693 & 0.2428 &0.7126 & 0.6559 & 0.5167 \\
\asrcomet{}$_\text{retrained}$  & 0.5618 & 0.4265 &0.4485 &0.5210 &0.2921 & 0.7489 &0.6123 &0.5159\\
\asrcomet{}$_\text{retrained with WMT}^\dagger$  & 0.5340 & 0.4348 & 0.5314 & 0.5659 & 0.2635 & 0.7308 & 0.6436 & 0.5291 \\
\blaser{}$_\text{s}$ & \textbf{0.5774$^*$} & \textbf{0.5347$^*$} & \textbf{0.6059$^*$} & \textbf{0.5730} & \textbf{0.3297$^*$} &\textbf{0.7512} & \textbf{0.7146$^*$} &\textbf{0.5838\phantom{*}}\\
\bottomrule
    \end{tabular}
    \caption{Pearson's correlation on the test set. Best results in bold. Results marked with $*$ pass the significance test with with $p\text{-value} < 0.05$ when compared against the baseline metric marked by $\dagger$ in the same category. 
    }
    \label{tab:results}
\end{table*}

\begin{table*}[t]
\small
    \centering
\sisetup{detect-weight,
         mode=text, 
         table-format = 1.4
         }
    \begin{tabular}{lSSSSSSSS}
    \toprule
  & \textbf{es$\rightarrow$en} & \textbf{ru$\rightarrow$en} & \textbf{hk$\rightarrow$en} & \textbf{fr$\rightarrow$en} & \textbf{en$\rightarrow$de} & \textbf{en$\rightarrow$es} & \textbf{en$\rightarrow$fr}  & \textbf{average}\\ \midrule  
Speech-only & 0.5774 & \bf{0.5347} & \bf{0.6059} & \bf{0.5730} & 0.3297 & \bf{0.7512} & \bf{0.7146} & \bf{0.5838} \\
Combined &  \bf{0.5791} & 0.5295 & 0.5988 & 0.5459 & \bf{0.3348} & 0.7456 & 0.7037 & 0.5767 \\
\bottomrule
    \end{tabular}
    \caption{Pearson's correlation on the test set. Best results are in bold. We compare \blasersup when training with speech data only and training with both speech and text data. For testing, we always evaluate models on speech data.}
    \label{tab:multimodal-supervision}
\end{table*}

\begin{table*}[t]
\small
    \centering
\sisetup{detect-weight,
         mode=text, 
         table-format = 1.4
         }
    \begin{tabular}{cccSSSSSSSSS}
    \toprule
  \multicolumn{3}{l}{\textbf{Modalities}}
  & \textbf{es$\rightarrow$en} & \textbf{ru$\rightarrow$en} & \textbf{hk$\rightarrow$en} & \textbf{fr$\rightarrow$en} & \textbf{en$\rightarrow$de} & \textbf{en$\rightarrow$es} & \textbf{en$\rightarrow$fr}  & \textbf{average}\\ \midrule  
(Speech,& Speech,& Speech) & \bf 0.5774 & \bf 0.5347 & \textbf{0.6059} &\bf 0.5730 & 0.3297 & \bf 0.7512 & \textbf{0.7146} &\textbf{0.5838} \\
(Speech, &Speech,& Text) & 0.5541 & 0.5164 & 0.5754 &	0.5425 & \bf 0.3675 & 0.7485 & 0.6688 & 0.5676 \\
(Speech,& Text, &Text) & 0.5460 &0.4866 &0.5616 &0.4741 &0.3393 &0.7372 &0.6285 &0.5390\\
(Text,& Text,& Text) & 0.4555 &0.4094 &0.5350 &0.4505 &0.2710 &0.6544 &0.5882 &0.4806 \\
\bottomrule
    \end{tabular}
    \caption{Pearson's correlation on the test set. Best results are in bold. $(x, y, z)$ indicates the modality used for source input ($x$), translation output ($y$), and reference ($z$). We train and evaluate \blasersup on the same modality combinations.
    }
    \label{tab:multimodal-results}
\end{table*}

\begin{table*}[h!]
\small
    \centering
\sisetup{detect-weight,
         mode=text, 
         table-format = 2.4
         }
    \begin{tabular}{cSSSSSSSS}
    \toprule
   
  & \textbf{es$\rightarrow$en} & \textbf{ru$\rightarrow$en} & \textbf{hk$\rightarrow$en} & \textbf{fr$\rightarrow$en} & \textbf{en$\rightarrow$de} & \textbf{en$\rightarrow$es} & \textbf{en$\rightarrow$fr}  & \textbf{average}\\ \midrule  
\asrsentbleu{} & 0.3226 & 0.1588 & 0.2863 & 0.3277 & 0.1259 & 0.4929 & 0.4393& 0.3076 \\
$\Delta$ & -0.0222 & -0.0244 & -0.0033 & -0.0161 & \bf -0.1161 & -0.0467 & -0.0341 & -0.0376 \\ 
\addlinespace[0.5em]
\asrchrf{} & 0.3910 & 0.2324 & 0.3356 & 0.3927 & 0.1673 & 0.6032 & 0.5177 & 0.3771 \\
$\Delta$ & -0.0195 & -0.0204 & 0.0017 & -0.0125 & \bf -0.1201 & -0.0757 & -0.0206 & -0.0382 \\ 
\addlinespace[0.5em]
\asrcomet{} & 0.5238 &0.3988 &0.5138 & 0.5693 & 0.2428 &0.7126 & 0.6559 & 0.5167 \\
$\Delta$ & -0.0164 & -0.0443 & -0.0602 & -0.0185 & \bf -0.0929 & -0.0281 & -0.0057 & -0.0380 \\ 
\bottomrule
    \end{tabular}
    \caption{Pearson's correlation on the test set. ``$\Delta$'' rows show the performance differences when using transcripts produced by ASR systems instead of humans for the source input and reference. Negative differences indicate performance drops. We highlight the results for en$\rightarrow$de as they are severely affected by the change.}
    \label{tab:asr-results}
\end{table*}

\subsection{Training and Evaluation}

\paragraph{\laser Encoders.} We use the speech \laser{} encoders released in \citet{duquenne2022speechmatrix} except for English and Hokkien.\footnote{\url{https://github.com/facebookresearch/fairseq/tree/ust/examples/speech_matrix}}. For Hokkien speech \laser{} encoder, we followed the training procedure presented in \cite{chen2022speech} using the same pretrained model and training data. For the English speech \laser{} encoder, we fine-tuned XLSR 2B \cite{babu2021xls} on several ASR datasets including CoVoST2 \cite{wang2021covost}, Common Voice \cite{DBLP:conf/lrec/ArdilaBDKMHMSTW20}, EuroparlST \cite{DBLP:conf/icassp/Iranzo-SanchezS20}, \mustc \cite{di-gangi-etal-2019-must}, Voxpopuli \cite{DBLP:conf/acl/WangRLWTHWPD20} and Librispeech \cite{panayotov2015librispeech}. 

\paragraph{Training Setup and Hyperparameters.} For \blasersup, the regressor has two hidden layers of sizes 3072 and 1536, similar to \comet{}. We keep the \laser encoders fixed during training. We use a learning rate of \num{5e-5} and employ learning rate annealing with a linear schedule. When training \comet, we follow the official implementation and fine-tune the entire model from the \textsc{xlm-r-large} model checkpoint \cite{conneau-etal-2020-unsupervised}. For both \blasersup and \comet, we train them for 20 epochs. We standardize the human ratings in our training set by subtracting them with a mean and a variance computed based on the entire training set.

\paragraph{Computational Cost.} We trained \blasersup using 1 Quadro GV100 and the training takes less than one hour. We used 4 Tesla V100 to train \comet{} and the training takes more than two days.

\paragraph{Evaluation.}
We compute Pearson's correlation at the sentence level between the automatic and human rating scores. Given that our test sets are relatively small, we perform statistical significance test using the bootstrap method from \citet{koehn-2004-statistical}.\footnote{\url{https://github.com/neubig/util-scripts/blob/master/paired-bootstrap.py}}

\section{Experimental Results and Analysis}

In this section we report the main results of our proposed metric \blaser{}, on two different settings (unsupervised and supervised) and we compare it to widely used baseline text-based metrics. 
Additionally, we report an analysis at various levels, including the impact of evaluating using different modalities and a qualitative inspection of several examples to observe scores of various metrics for particular examples.

\subsection{Main Results}
\label{sec:results}

We report unsupervised and supervised results in Table~\ref{tab:results}. We note that results that fail to pass the significance test are neither better nor worse significantly than the corresponding baseline.

Generally, model-based metrics perform significantly better than string-based ones. Among the unsupervised metrics, \blaseruns{} performance improves significantly over \asrsentbleu{} and \asrchrf{} for all language directions except for en$\rightarrow$es, showing the capabilities of \blaser in capturing semantic information even when human annotations are absent.

Among the supervised metrics, we see that \blasersup almost always performs better than the official \asrbertscore{} and \asrcomet{}. When compared to the stronger baseline \asrcomet{}$_\text{retrained with WMT}$, \blasersup{} is better than the baseline significantly in four language directions and they are comparable in the other three directions.

We also find that \blaser can generalize training signal to languages where there is no training data available. Specifically, if we compare \blasersup{} to \blaseruns{}, we see that \blasersup always improves over the unsupervised version. Also, for the language directions where there is no training data (i.e., hk$\rightarrow$en, fr$\rightarrow$en, en$\rightarrow$fr), \blasersup{} still beats \blaseruns. Additionally, we observe that hk$\rightarrow$en and ru$\rightarrow$en are two of the language directions for which \blasersup shows significant improvements over \asrcomet{}, confirming the zero-shot capabilities of our proposed methods in comparison to existing metrics. 

\subsection{Analysis}

\paragraph{Cross-Modal Data.}
Considering that \laser can conveniently encode text and speech data into a shared embedding space, we conduct experiments involving both text and speech data with the text encoders from \citet{heffernan2022bitext} for \blasersup{}.
In particular, we embed the source input, translation output, and reference using either the speech or text \laser encoders. That is, a data instance formed by embeddings from speech data will result in four instances in this new setting due to different modality combinations. We then evaluate the models on the speech data in our test set. The results in Table \ref{tab:multimodal-supervision} show that combining supervision from different modalities does not help improve model performance. It is likely because the embedding space is shared between text and speech and therefore adding textual embeddings do not provide extra information.

\paragraph{Cross-Modal Supervision.} We also look into the benefits of leveraging speech embeddings by comparing several supervised configurations for \blasersup. We report these results in Table~\ref{tab:multimodal-results} where we experiment with different modality combinations during training and testing. The results show that the best results on average are the ones using speech modality for the source input, translation output, and reference. Interestingly, every time that we replace speech with text in the modality combinations, we see performance drops. We find that replacing reference \audio with text leads to the slightest performance drop, which is likely due to the fact that they are synthesized and thus do not provide extra information than text. We also find that replacing speech data with text for the source input and translation output makes \blasersup similar or even worse than \asrcomet{}$_\text{retrained with WMT}$, confirming the benefits of using speech data for evaluation S2ST systems. 
Additionally, in Appendix \ref{sec:appendix-analysis} we evaluate \blasersup on different modality combinations when training on speech data only.

\paragraph{Impact of Human-Written vs ASR-transcriptions.}
To investigate the impact of using transcripts generated by ASR systems rather than human-written inputs and references, we replace the human-written source input and reference with the ones generated by ASR systems. We note that in this case, all the transcripts are obtained via ASR systems, simulating an evaluation setting where only audio data is available. We show the results in Table~\ref{tab:asr-results} where we find that the human-written transcripts are less helpful on those to-English language directions than the from-English ones. We hypothesize that this is in part due to the quality of ASR systems as these ASR-based metrics depend more on references than source inputs and English ASR systems tend to be of better quality than the non-English ones \cite{khare21_interspeech}.

\paragraph{Impact of Using Source and Reference.}

\begin{table*}[t]
\small
    \centering
\sisetup{detect-weight,
         mode=text, 
         table-format = 1.4
         }
    \begin{tabular}{cSSSSSSSS}
    \toprule
  & \textbf{es$\rightarrow$en} & \textbf{ru$\rightarrow$en} & \textbf{hk$\rightarrow$en} & \textbf{fr$\rightarrow$en} & \textbf{en$\rightarrow$de} & \textbf{en$\rightarrow$es} & \textbf{en$\rightarrow$fr}  & \textbf{average}\\ \midrule  
$\text{cos} (h_\text{ref}, h_\text{mt})+\text{cos}(h_\text{src}, h_\text{mt})$ & \bf 0.4970 & \bf 0.4326 & \bf 0.4940 & \bf 0.4744 & \bf 0.3148 & 0.5843 & \bf 0.6356 & \bf 0.4904 \\
$\text{cos} (h_\text{ref}, h_\text{mt})$ & 0.4392 & 0.2855 & 0.4051 & 0.4144 & 0.1388 & 0.4516 & 0.5588 & 0.3848 \\
$\text{cos}(h_\text{src}, h_\text{mt})$& 0.4392 & 0.4182 & 0.4723 & 0.4450 & 0.2654 & \bf 0.6411 & 0.6215 & 0.4718 \\
\bottomrule
    \end{tabular}
    \caption{Pearson's correlation on the test set. Best results are in bold. We evaluate the contributions of two individual terms in \blaseruns (Equation~\ref{eq:1}) to the final performance.}
    \label{tab:impact-of-src-ref}
\end{table*}

We investigate the impact of using source and reference \audios when computing \blaser scores. We evaluate this impact on \blaseruns by reporting the performance of individual terms in Equation~\ref{eq:1}. See the results in Table~\ref{tab:impact-of-src-ref}. In general, we find the source input generates better correlations with human ratings than reference. Combining the two leads to the best performance.


\paragraph{Qualitative Analysis.}

To get a sense of the qualitative differences between \blaser{} and text-based scores, and better understand what kind of nuances are captured, we manually inspect sample sentences. A selection is presented in Table \ref{tab:qual_analysis}. In each of these examples, the text and generated audio perfectly match, discarding any influence potentially introduced by the ASR model.
In cases where the output vocabulary does not perfectly match the reference but is still valid, \blaser{} seems able to capture the semantics and produce a meaningful score. In the first example, \asrsentbleu{} is very much impacted by the vocabulary mismatch, while \blaser{} and \asrcomet{} yield high scores, in line with human evaluation.
\blaser{} also seem to detect clear mistranslations better than either of \asrcomet{} or \asrsentbleu{}. In the second example, the end of the output sentence makes little sense. Only \blaser{} accounts for this properly and produces a score aligned with human judgment. In the third example, \asrcomet{} returns a high score despite the mistranslated verb which heavily changes the meaning of the sentence.

\begin{table*}[h!]
    \centering\small
    \begin{tabular}{|p{0.24\textwidth}|p{0.24\textwidth}|p{0.24\textwidth}|c|c|c|c|}\hline
      \multicolumn{1}{|c|}{source input}  &  \multicolumn{1}{c|}{translation output} & \multicolumn{1}{c|}{reference} & HR & BR & CT & BU \\\hline\hline
        The pollution in Santiago, which is one of the most polluted capitals historically in Latin America, has dropped substantially. & die verschmutzung in santiago einem der am stärksten verschmutzten hauptstädte latein-amerikas ist erheblich gesungen (the pollution in santiago one the at strongest polluted capital cities latin america is significantly sung) & Die Umweltverschmutzung in Santiago, das historisch gesehen eine der Städte mit der höchsten Umweltverschmutzung in ganz Lateinamerika ist, ist viel geringer geworden.  & 4.5 & 0.2 & 0.9 & 4.0 \\
        \hline 
        And for those of us that are in the know, we know that's text-speak, or SMS language. & diejenigen von uns die das kennen wissen das ist zum spracher (those from us the the know to know the is for the speaker) & Diejenigen von uns, die das kennen, wissen: Das ist SMS-Sprache. & 2.5 & 0.0 & 0.9 & 78.6 \\ 
        \hline 
        So, when I say, "Oh, Aaron is..." It's because Aaron still is.  & wenn ich aron sehe liegt das daran dass aron es immer noch ist (if I aron see located the to it that aron it always still is)  & Wenn ich also sage: „Oh, Aaron ist ...“, dann sage ich das, weil Aaron immer noch ist. & 3.5 & -0.1 & 0.9 & 12.9 \\ 
        \hline 
    \end{tabular}
    \caption{The examples from the en$\rightarrow$de test set and the corresponding scores given by different metrics.
    HR=Human Ratings. BR=\blasersup{}. CT=\asrcomet{}. BU=\asrsentbleu{}. Sentences in parenthesis are gloss for translation outputs.}
    \label{tab:qual_analysis}
\end{table*}

\section{Conclusion and Future Work}

We have introduced \blaser{}, a text-free metric to evaluate speech-to-speech translation, which avoids the dependency on ASR models required by popular text-based metrics currently used in S2ST. We explored \blaser{} in both unsupervised and supervised settings. Experimental results in seven language directions show that \blaser{} outperforms or is comparable to strong text-based metrics in terms of correlation with human scores at the sentence-level. Moreover, our metric is effective in zero-shot scenarios. 

As for future work, we want to explore the use of speech references generated by humans and the impact of synthesized references. We also want to evaluate \blaser at the system-level with a much larger number of S2ST systems, and explore different approaches to aggregate the sentence-level scores from \blaser{} and we want to explore different speech and text representations as alternative to \laser{}.



\section*{Acknowledgment}

The authors would like to specially thank Ilia Kulikov, Sravya Popuri, Peng-Jen Chen, Changhan Wang and Kevin Heffernan for the valuable technical support and interesting discussions.

\bibliography{anthology,custom,blaser}

\begin{thebibliography}{74}
\expandafter\ifx\csname natexlab\endcsname\relax\def\natexlab#1{#1}\fi

\bibitem[{Anastasopoulos et~al.(2022)Anastasopoulos, Barrault, Bentivogli,
  Zanon~Boito, Bojar, Cattoni, Currey, Dinu, Duh, Elbayad, Emmanuel,
  Est{\`e}ve, Federico, Federmann, Gahbiche, Gong, Grundkiewicz, Haddow, Hsu,
  Javorsk{\'y}, Kloudov{\'a}, Lakew, Ma, Mathur, McNamee, Murray,
  N{\v{a}}dejde, Nakamura, Negri, Niehues, Niu, Ortega, Pino, Salesky, Shi,
  Sperber, St{\"u}ker, Sudoh, Turchi, Virkar, Waibel, Wang, and
  Watanabe}]{anastasopoulos-etal-2022-findings}
Antonios Anastasopoulos, Lo{\"\i}c Barrault, Luisa Bentivogli, Marcely
  Zanon~Boito, Ond{\v{r}}ej Bojar, Roldano Cattoni, Anna Currey, Georgiana
  Dinu, Kevin Duh, Maha Elbayad, Clara Emmanuel, Yannick Est{\`e}ve, Marcello
  Federico, Christian Federmann, Souhir Gahbiche, Hongyu Gong, Roman
  Grundkiewicz, Barry Haddow, Benjamin Hsu, D{\'a}vid Javorsk{\'y}, V{\u{e}}ra
  Kloudov{\'a}, Surafel Lakew, Xutai Ma, Prashant Mathur, Paul McNamee, Kenton
  Murray, Maria N{\v{a}}dejde, Satoshi Nakamura, Matteo Negri, Jan Niehues,
  Xing Niu, John Ortega, Juan Pino, Elizabeth Salesky, Jiatong Shi, Matthias
  Sperber, Sebastian St{\"u}ker, Katsuhito Sudoh, Marco Turchi, Yogesh Virkar,
  Alexander Waibel, Changhan Wang, and Shinji Watanabe. 2022.
\newblock \href {https://doi.org/10.18653/v1/2022.iwslt-1.10} {Findings of the
  {IWSLT} 2022 evaluation campaign}.
\newblock In \emph{Proceedings of the 19th International Conference on Spoken
  Language Translation (IWSLT 2022)}, pages 98--157, Dublin, Ireland (in-person
  and online). Association for Computational Linguistics.

\bibitem[{Ardila et~al.(2020)Ardila, Branson, Davis, Kohler, Meyer, Henretty,
  Morais, Saunders, Tyers, and Weber}]{DBLP:conf/lrec/ArdilaBDKMHMSTW20}
Rosana Ardila, Megan Branson, Kelly Davis, Michael Kohler, Josh Meyer, Michael
  Henretty, Reuben Morais, Lindsay Saunders, Francis~M. Tyers, and Gregor
  Weber. 2020.
\newblock Common voice: {A} massively-multilingual speech corpus.
\newblock In \emph{Proceedings of The 12th Language Resources and Evaluation
  Conference, {LREC} 2020, Marseille, France, May 11-16, 2020}, pages
  4218--4222. European Language Resources Association.

\bibitem[{Artetxe and Schwenk(2019)}]{artetxe-schwenk-2019-massively}
Mikel Artetxe and Holger Schwenk. 2019.
\newblock \href {https://doi.org/10.1162/tacl_a_00288} {Massively multilingual
  sentence embeddings for zero-shot cross-lingual transfer and beyond}.
\newblock \emph{Transactions of the Association for Computational Linguistics},
  7:597--610.

\bibitem[{Babu et~al.(2021)Babu, Wang, Tjandra, Lakhotia, Xu, Goyal, Singh, von
  Platen, Saraf, Pino et~al.}]{babu2021xls}
Arun Babu, Changhan Wang, Andros Tjandra, Kushal Lakhotia, Qiantong Xu, Naman
  Goyal, Kritika Singh, Patrick von Platen, Yatharth Saraf, Juan Pino, et~al.
  2021.
\newblock Xls-r: Self-supervised cross-lingual speech representation learning
  at scale.
\newblock \emph{arXiv preprint arXiv:2111.09296}.

\bibitem[{Baevski et~al.(2020{\natexlab{a}})Baevski, Zhou, Mohamed, and
  Auli}]{baevski2020wav2vec}
Alexei Baevski, Yuhao Zhou, Abdelrahman Mohamed, and Michael Auli.
  2020{\natexlab{a}}.
\newblock wav2vec 2.0: A framework for self-supervised learning of speech
  representations.
\newblock \emph{Advances in Neural Information Processing Systems},
  33:12449--12460.

\bibitem[{Baevski et~al.(2020{\natexlab{b}})Baevski, Zhou, Mohamed, and
  Auli}]{NEURIPS2020_92d1e1eb}
Alexei Baevski, Yuhao Zhou, Abdelrahman Mohamed, and Michael Auli.
  2020{\natexlab{b}}.
\newblock \href
  {https://proceedings.neurips.cc/paper/2020/file/92d1e1eb1cd6f9fba3227870bb6d7f07-Paper.pdf}
  {wav2vec 2.0: A framework for self-supervised learning of speech
  representations}.
\newblock In \emph{Advances in Neural Information Processing Systems},
  volume~33, pages 12449--12460. Curran Associates, Inc.

\bibitem[{Bapna et~al.(2022)Bapna, Cherry, Zhang, Jia, Johnson, Cheng, Khanuja,
  Riesa, and Conneau}]{bapna2022mslam}
Ankur Bapna, Colin Cherry, Yu~Zhang, Ye~Jia, Melvin Johnson, Yong Cheng, Simran
  Khanuja, Jason Riesa, and Alexis Conneau. 2022.
\newblock mslam: Massively multilingual joint pre-training for speech and text.
\newblock \emph{arXiv preprint arXiv:2202.01374}.

\bibitem[{Besacier et~al.(2022)Besacier, Ribeiro, Galibert, and
  Calapodescu}]{besacier2022textless}
Laurent Besacier, Swen Ribeiro, Olivier Galibert, and Ioan Calapodescu. 2022.
\newblock A textless metric for speech-to-speech comparison.
\newblock \emph{arXiv preprint arXiv:2210.11835}.

\bibitem[{Bińkowski et~al.(2020)Bińkowski, Donahue, Dieleman, Clark, Elsen,
  Casagrande, Cobo, and Simonyan}]{Bińkowski2020High}
Mikołaj Bińkowski, Jeff Donahue, Sander Dieleman, Aidan Clark, Erich Elsen,
  Norman Casagrande, Luis~C. Cobo, and Karen Simonyan. 2020.
\newblock \href {https://openreview.net/forum?id=r1gfQgSFDr} {High fidelity
  speech synthesis with adversarial networks}.
\newblock In \emph{International Conference on Learning Representations}.

\bibitem[{Bojar et~al.(2017)Bojar, Graham, and
  Kamran}]{bojar-etal-2017-results}
Ond{\v{r}}ej Bojar, Yvette Graham, and Amir Kamran. 2017.
\newblock \href {https://doi.org/10.18653/v1/W17-4755} {Results of the {WMT}17
  metrics shared task}.
\newblock In \emph{Proceedings of the Second Conference on Machine
  Translation}, pages 489--513, Copenhagen, Denmark. Association for
  Computational Linguistics.

\bibitem[{Bojar et~al.(2016)Bojar, Graham, Kamran, and
  Stanojevi{\'c}}]{bojar-etal-2016-results}
Ond{\v{r}}ej Bojar, Yvette Graham, Amir Kamran, and Milo{\v{s}} Stanojevi{\'c}.
  2016.
\newblock \href {https://doi.org/10.18653/v1/W16-2302} {Results of the {WMT}16
  metrics shared task}.
\newblock In \emph{Proceedings of the First Conference on Machine Translation:
  Volume 2, Shared Task Papers}, pages 199--231, Berlin, Germany. Association
  for Computational Linguistics.

\bibitem[{Chen et~al.(2022{\natexlab{a}})Chen, Tran, Yang, Du, Kao, Chung,
  Tomasello, Duquenne, Schwenk, Gong et~al.}]{chen2022speech}
Peng-Jen Chen, Kevin Tran, Yilin Yang, Jingfei Du, Justine Kao, Yu-An Chung,
  Paden Tomasello, Paul-Ambroise Duquenne, Holger Schwenk, Hongyu Gong, et~al.
  2022{\natexlab{a}}.
\newblock Speech-to-speech translation for a real-world unwritten language.
\newblock \emph{arXiv preprint arXiv:2211.06474}.

\bibitem[{Chen et~al.(2022{\natexlab{b}})Chen, Zhang, Rosenberg, Ramabhadran,
  Moreno, Bapna, and Zen}]{chen2022maestro}
Zhehuai Chen, Yu~Zhang, Andrew Rosenberg, Bhuvana Ramabhadran, Pedro Moreno,
  Ankur Bapna, and Heiga Zen. 2022{\natexlab{b}}.
\newblock Maestro: Matched speech text representations through modality
  matching.
\newblock \emph{arXiv preprint arXiv:2204.03409}.

\bibitem[{Conneau et~al.(2020{\natexlab{a}})Conneau, Baevski, Collobert,
  Mohamed, and Auli}]{conneau2020unsupervised}
Alexis Conneau, Alexei Baevski, Ronan Collobert, Abdelrahman Mohamed, and
  Michael Auli. 2020{\natexlab{a}}.
\newblock Unsupervised cross-lingual representation learning for speech
  recognition.
\newblock \emph{arXiv preprint arXiv:2006.13979}.

\bibitem[{Conneau et~al.(2020{\natexlab{b}})Conneau, Khandelwal, Goyal,
  Chaudhary, Wenzek, Guzm{\'a}n, Grave, Ott, Zettlemoyer, and
  Stoyanov}]{conneau-etal-2020-unsupervised}
Alexis Conneau, Kartikay Khandelwal, Naman Goyal, Vishrav Chaudhary, Guillaume
  Wenzek, Francisco Guzm{\'a}n, Edouard Grave, Myle Ott, Luke Zettlemoyer, and
  Veselin Stoyanov. 2020{\natexlab{b}}.
\newblock \href {https://doi.org/10.18653/v1/2020.acl-main.747} {Unsupervised
  cross-lingual representation learning at scale}.
\newblock In \emph{Proceedings of the 58th Annual Meeting of the Association
  for Computational Linguistics}, pages 8440--8451, Online. Association for
  Computational Linguistics.

\bibitem[{Denkowski and Lavie(2014)}]{denkowski-lavie-2014-meteor}
Michael Denkowski and Alon Lavie. 2014.
\newblock \href {https://doi.org/10.3115/v1/W14-3348} {Meteor universal:
  Language specific translation evaluation for any target language}.
\newblock In \emph{Proceedings of the Ninth Workshop on Statistical Machine
  Translation}, pages 376--380, Baltimore, Maryland, USA. Association for
  Computational Linguistics.

\bibitem[{Di~Gangi et~al.(2019)Di~Gangi, Cattoni, Bentivogli, Negri, and
  Turchi}]{di-gangi-etal-2019-must}
Mattia~A. Di~Gangi, Roldano Cattoni, Luisa Bentivogli, Matteo Negri, and Marco
  Turchi. 2019.
\newblock \href {https://doi.org/10.18653/v1/N19-1202} {{M}u{ST}-{C}: a
  {M}ultilingual {S}peech {T}ranslation {C}orpus}.
\newblock In \emph{Proceedings of the 2019 Conference of the North {A}merican
  Chapter of the Association for Computational Linguistics: Human Language
  Technologies, Volume 1 (Long and Short Papers)}, pages 2012--2017,
  Minneapolis, Minnesota. Association for Computational Linguistics.

\bibitem[{Duquenne et~al.(2022)Duquenne, Gong, Dong, Du, Lee, Goswani, Wang,
  Pino, Sagot, and Schwenk}]{duquenne2022speechmatrix}
Paul-Ambroise Duquenne, Hongyu Gong, Ning Dong, Jingfei Du, Ann Lee, Vedanuj
  Goswani, Changhan Wang, Juan Pino, Beno{\^\i}t Sagot, and Holger Schwenk.
  2022.
\newblock Speechmatrix: A large-scale mined corpus of multilingual
  speech-to-speech translations.
\newblock \emph{arXiv preprint arXiv:2211.04508}.

\bibitem[{Duquenne et~al.(2021)Duquenne, Gong, and
  Schwenk}]{NEURIPS2021_8466f9ac}
Paul-Ambroise Duquenne, Hongyu Gong, and Holger Schwenk. 2021.
\newblock \href
  {https://proceedings.neurips.cc/paper/2021/file/8466f9ace6a9acbe71f75762ffc890f1-Paper.pdf}
  {Multimodal and multilingual embeddings for large-scale speech mining}.
\newblock In \emph{Advances in Neural Information Processing Systems},
  volume~34, pages 15748--15761. Curran Associates, Inc.

\bibitem[{Feng et~al.(2022)Feng, Yang, Cer, Arivazhagan, and
  Wang}]{feng-etal-2022-language}
Fangxiaoyu Feng, Yinfei Yang, Daniel Cer, Naveen Arivazhagan, and Wei Wang.
  2022.
\newblock \href {https://doi.org/10.18653/v1/2022.acl-long.62}
  {Language-agnostic {BERT} sentence embedding}.
\newblock In \emph{Proceedings of the 60th Annual Meeting of the Association
  for Computational Linguistics (Volume 1: Long Papers)}, pages 878--891,
  Dublin, Ireland. Association for Computational Linguistics.

\bibitem[{Freitag et~al.(2021)Freitag, Rei, Mathur, Lo, Stewart, Foster, Lavie,
  and Bojar}]{freitag-etal-2021-results}
Markus Freitag, Ricardo Rei, Nitika Mathur, Chi-kiu Lo, Craig Stewart, George
  Foster, Alon Lavie, and Ond{\v{r}}ej Bojar. 2021.
\newblock \href {https://aclanthology.org/2021.wmt-1.73} {Results of the
  {WMT}21 metrics shared task: Evaluating metrics with expert-based human
  evaluations on {TED} and news domain}.
\newblock In \emph{Proceedings of the Sixth Conference on Machine Translation},
  pages 733--774, Online. Association for Computational Linguistics.

\bibitem[{Graves et~al.(2006)Graves, Fern\'{a}ndez, Gomez, and
  Schmidhuber}]{10.1145/1143844.1143891}
Alex Graves, Santiago Fern\'{a}ndez, Faustino Gomez, and J\"{u}rgen
  Schmidhuber. 2006.
\newblock \href {https://doi.org/10.1145/1143844.1143891} {Connectionist
  temporal classification: Labelling unsegmented sequence data with recurrent
  neural networks}.
\newblock In \emph{Proceedings of the 23rd International Conference on Machine
  Learning}, ICML '06, page 369–376, New York, NY, USA. Association for
  Computing Machinery.

\bibitem[{Grosman(2021{\natexlab{a}})}]{grosman2021voxpopuli-fr-wav2vec2-large-french}
Jonatas Grosman. 2021{\natexlab{a}}.
\newblock Fine-tuned {F}rench {V}oxpopuli wav2vec2 large model for speech
  recognition in {F}rench.
\newblock
  \url{https://huggingface.co/jonatasgrosman/wav2vec2-large-fr-voxpopuli-french}.

\bibitem[{Grosman(2021{\natexlab{b}})}]{grosman2021xlsr53-large-spanish}
Jonatas Grosman. 2021{\natexlab{b}}.
\newblock Fine-tuned {XLSR}-53 large model for speech recognition in {S}panish.
\newblock
  \url{https://huggingface.co/jonatasgrosman/wav2vec2-large-xlsr-53-spanish}.

\bibitem[{Grosman(2022)}]{grosman2021xlsr-1b-german}
Jonatas Grosman. 2022.
\newblock Fine-tuned {XLS-R} 1{B} model for speech recognition in {G}erman.
\newblock \url{https://huggingface.co/jonatasgrosman/wav2vec2-xls-r-1b-german}.

\bibitem[{Heffernan et~al.(2022)Heffernan, {\c{C}}elebi, and
  Schwenk}]{heffernan2022bitext}
Kevin Heffernan, Onur {\c{C}}elebi, and Holger Schwenk. 2022.
\newblock Bitext mining using distilled sentence representations for
  low-resource languages.
\newblock \emph{arXiv preprint arXiv:2205.12654}.

\bibitem[{Hsu et~al.(2021)Hsu, Bolte, Tsai, Lakhotia, Salakhutdinov, and
  Mohamed}]{hsu2021hubert}
Wei-Ning Hsu, Benjamin Bolte, Yao-Hung~Hubert Tsai, Kushal Lakhotia, Ruslan
  Salakhutdinov, and Abdelrahman Mohamed. 2021.
\newblock Hubert: Self-supervised speech representation learning by masked
  prediction of hidden units.
\newblock \emph{IEEE/ACM Transactions on Audio, Speech, and Language
  Processing}, 29:3451--3460.

\bibitem[{Inaguma et~al.(2022)Inaguma, Popuri, Kulikov, Chen, Wang, Tang, Lee,
  Watanabe, and Pino}]{inaguma2022unity}
Hirofumi Inaguma, Sravya Popuri, Ilia Kulikov, Peng-Jen Chen, Changhan Wang,
  Yun Tang, Ann Lee, Shinji Watanabe, and Juan Pino. 2022.
\newblock Unity: Two-pass direct speech-to-speech translation with discrete
  units.
\newblock \emph{arXiv preprint}.

\bibitem[{Iranzo-S{\'a}nchez et~al.(2022)Iranzo-S{\'a}nchez, Jorge~Cano,
  P{\'e}rez-Gonz{\'a}lez-de Martos, Gim{\'e}nez~Pastor, Garc{\'e}s
  D{\'\i}az-Mun{\'\i}o, Baquero-Arnal, Silvestre-Cerd{\`a}, Civera~Saiz,
  Sanchis, and Juan}]{iranzo-sanchez-etal-2022-mllp}
Javier Iranzo-S{\'a}nchez, Javier Jorge~Cano, Alejandro
  P{\'e}rez-Gonz{\'a}lez-de Martos, Adri{\'a}n Gim{\'e}nez~Pastor, Gon{\c{c}}al
  Garc{\'e}s D{\'\i}az-Mun{\'\i}o, Pau Baquero-Arnal, Joan~Albert
  Silvestre-Cerd{\`a}, Jorge Civera~Saiz, Albert Sanchis, and Alfons Juan.
  2022.
\newblock \href {https://doi.org/10.18653/v1/2022.iwslt-1.22} {{MLLP}-{VRAIN}
  {UPV} systems for the {IWSLT} 2022 simultaneous speech translation and
  speech-to-speech translation tasks}.
\newblock In \emph{Proceedings of the 19th International Conference on Spoken
  Language Translation (IWSLT 2022)}, pages 255--264, Dublin, Ireland
  (in-person and online). Association for Computational Linguistics.

\bibitem[{Iranzo{-}S{\'{a}}nchez et~al.(2020)Iranzo{-}S{\'{a}}nchez,
  Silvestre{-}Cerd{\`{a}}, Jorge, Rosell{\'{o}}, Gim{\'{e}}nez, Sanch{\'{\i}}s,
  Civera, and Juan}]{DBLP:conf/icassp/Iranzo-SanchezS20}
Javier Iranzo{-}S{\'{a}}nchez, Joan~Albert Silvestre{-}Cerd{\`{a}}, Javier
  Jorge, Nahuel Rosell{\'{o}}, Adri{\`{a}} Gim{\'{e}}nez, Albert
  Sanch{\'{\i}}s, Jorge Civera, and Alfons Juan. 2020.
\newblock Europarl-st: {A} multilingual corpus for speech translation of
  parliamentary debates.
\newblock In \emph{2020 {IEEE} International Conference on Acoustics, Speech
  and Signal Processing, {ICASSP} 2020, Barcelona, Spain, May 4-8, 2020}, pages
  8229--8233. {IEEE}.

\bibitem[{Javed et~al.(2022)Javed, Doddapaneni, Raman, Bhogale, Ramesh,
  Kunchukuttan, Kumar, and Khapra}]{DBLP:journals/corr/abs-2111-03945}
Tahir Javed, Sumanth Doddapaneni, Abhigyan Raman, Kaushal~Santosh Bhogale,
  Gowtham Ramesh, Anoop Kunchukuttan, Pratyush Kumar, and Mitesh~M. Khapra.
  2022.
\newblock Towards building {ASR} systems for the next billion users.
\newblock In \emph{Proceedings of AAAI}.

\bibitem[{Jia et~al.(2019)Jia, Weiss, Biadsy, Macherey, Johnson, Chen, and
  Wu}]{Jia2019DirectST}
Ye~Jia, Ron~J. Weiss, Fadi Biadsy, Wolfgang Macherey, Melvin Johnson, Z.~Chen,
  and Yonghui Wu. 2019.
\newblock Direct speech-to-speech translation with a sequence-to-sequence
  model.
\newblock In \emph{INTERSPEECH}.

\bibitem[{Khare et~al.(2021)Khare, Mittal, Diwan, Sarawagi, Jyothi, and
  Bharadwaj}]{khare21_interspeech}
Shreya Khare, Ashish Mittal, Anuj Diwan, Sunita Sarawagi, Preethi Jyothi, and
  Samarth Bharadwaj. 2021.
\newblock \href {https://doi.org/10.21437/Interspeech.2021-2062} {{Low Resource
  ASR: The Surprising Effectiveness of High Resource Transliteration}}.
\newblock In \emph{Proc. Interspeech 2021}, pages 1529--1533.

\bibitem[{Khurana et~al.(2022)Khurana, Laurent, and Glass}]{khurana2022samu}
Sameer Khurana, Antoine Laurent, and James Glass. 2022.
\newblock Samu-xlsr: Semantically-aligned multimodal utterance-level
  cross-lingual speech representation.
\newblock \emph{arXiv preprint arXiv:2205.08180}.

\bibitem[{Kocmi et~al.(2021)Kocmi, Federmann, Grundkiewicz, Junczys-Dowmunt,
  Matsushita, and Menezes}]{kocmi-etal-2021-ship}
Tom Kocmi, Christian Federmann, Roman Grundkiewicz, Marcin Junczys-Dowmunt,
  Hitokazu Matsushita, and Arul Menezes. 2021.
\newblock \href {https://aclanthology.org/2021.wmt-1.57} {To ship or not to
  ship: An extensive evaluation of automatic metrics for machine translation}.
\newblock In \emph{Proceedings of the Sixth Conference on Machine Translation},
  pages 478--494, Online. Association for Computational Linguistics.

\bibitem[{Koehn(2004)}]{koehn-2004-statistical}
Philipp Koehn. 2004.
\newblock \href {https://aclanthology.org/W04-3250} {Statistical significance
  tests for machine translation evaluation}.
\newblock In \emph{Proceedings of the 2004 Conference on Empirical Methods in
  Natural Language Processing}, pages 388--395, Barcelona, Spain. Association
  for Computational Linguistics.

\bibitem[{Kominek et~al.(2008)Kominek, Schultz, and
  Black}]{kominek2008synthesizer}
John Kominek, Tanja Schultz, and Alan~W Black. 2008.
\newblock Synthesizer voice quality of new languages calibrated with mean mel
  cepstral distortion.
\newblock In \emph{SLTU}, pages 63--68.

\bibitem[{Lavie et~al.(1997)Lavie, Waibel, Levin, Finke, Gates, Gavalda,
  Zeppenfeld, and Zhan}]{lavie}
Alon Lavie, A.~Waibel, Lori Levin, M.~Finke, Donna Gates, Marsal Gavalda,
  Torsten Zeppenfeld, and Puming Zhan. 1997.
\newblock \href {https://doi.org/10.1109/ICASSP.1997.599557} {Janus-iii:
  speech-to-speech translation in multiple languages}.
\newblock pages 99 -- 102 vol.1.

\bibitem[{Lazzari(2006)}]{lazzari-2006-tc}
Gianni Lazzari. 2006.
\newblock \href {https://aclanthology.org/2006.iwslt-plenaries.1} {{TC}-{STAR}:
  a speech to speech translation project}.
\newblock In \emph{Proceedings of the Third International Workshop on Spoken
  Language Translation: Plenaries}, Kyoto, Japan.

\bibitem[{Lee et~al.(2022{\natexlab{a}})Lee, Chen, Wang, Gu, Popuri, Ma,
  Polyak, Adi, He, Tang, Pino, and Hsu}]{lee-etal-2022-direct}
Ann Lee, Peng-Jen Chen, Changhan Wang, Jiatao Gu, Sravya Popuri, Xutai Ma, Adam
  Polyak, Yossi Adi, Qing He, Yun Tang, Juan Pino, and Wei-Ning Hsu.
  2022{\natexlab{a}}.
\newblock \href {https://doi.org/10.18653/v1/2022.acl-long.235} {Direct
  speech-to-speech translation with discrete units}.
\newblock In \emph{Proceedings of the 60th Annual Meeting of the Association
  for Computational Linguistics (Volume 1: Long Papers)}, pages 3327--3339,
  Dublin, Ireland. Association for Computational Linguistics.

\bibitem[{Lee et~al.(2022{\natexlab{b}})Lee, Gong, Duquenne, Schwenk, Chen,
  Wang, Popuri, Adi, Pino, Gu, and Hsu}]{lee-etal-2022-textless}
Ann Lee, Hongyu Gong, Paul-Ambroise Duquenne, Holger Schwenk, Peng-Jen Chen,
  Changhan Wang, Sravya Popuri, Yossi Adi, Juan Pino, Jiatao Gu, and Wei-Ning
  Hsu. 2022{\natexlab{b}}.
\newblock \href {https://doi.org/10.18653/v1/2022.naacl-main.63} {Textless
  speech-to-speech translation on real data}.
\newblock In \emph{Proceedings of the 2022 Conference of the North American
  Chapter of the Association for Computational Linguistics: Human Language
  Technologies}, pages 860--872, Seattle, United States. Association for
  Computational Linguistics.

\bibitem[{Liao et~al.(2020)Liao, Chang, Tiun, Su, Khoo, Tsay, Tan, Kang,
  Thiann, Iunn, Yang, and Liang}]{9295019}
Yuan-Fu Liao, Chia-Yu Chang, Hak-Khiam Tiun, Huang-Lan Su, Hui-Lu Khoo, Jane~S.
  Tsay, Le-Kun Tan, Peter Kang, Tsun-guan Thiann, Un-Gian Iunn, Jyh-Her Yang,
  and Chih-Neng Liang. 2020.
\newblock \href {https://doi.org/10.1109/O-COCOSDA50338.2020.9295019} {Formosa
  speech recognition challenge 2020 and taiwanese across taiwan corpus}.
\newblock In \emph{2020 23rd Conference of the Oriental COCOSDA International
  Committee for the Co-ordination and Standardisation of Speech Databases and
  Assessment Techniques (O-COCOSDA)}, pages 65--70.

\bibitem[{Licht et~al.(2022)Licht, Gao, Lam, Guzman, Diab, and
  Koehn}]{licht-etal-2022-consistent}
Daniel Licht, Cynthia Gao, Janice Lam, Francisco Guzman, Mona Diab, and Philipp
  Koehn. 2022.
\newblock \href {https://aclanthology.org/2022.amta-research.24} {Consistent
  human evaluation of machine translation across language pairs}.
\newblock In \emph{Proceedings of the 15th biennial conference of the
  Association for Machine Translation in the Americas (Volume 1: Research
  Track)}, pages 309--321, Orlando, USA. Association for Machine Translation in
  the Americas.

\bibitem[{Liu et~al.(2020)Liu, Gu, Goyal, Li, Edunov, Ghazvininejad, Lewis, and
  Zettlemoyer}]{liu-etal-2020-multilingual-denoising}
Yinhan Liu, Jiatao Gu, Naman Goyal, Xian Li, Sergey Edunov, Marjan
  Ghazvininejad, Mike Lewis, and Luke Zettlemoyer. 2020.
\newblock \href {https://doi.org/10.1162/tacl_a_00343} {Multilingual denoising
  pre-training for neural machine translation}.
\newblock \emph{Transactions of the Association for Computational Linguistics},
  8:726--742.

\bibitem[{Ma et~al.(2018)Ma, Bojar, and Graham}]{ma-etal-2018-results}
Qingsong Ma, Ond{\v{r}}ej Bojar, and Yvette Graham. 2018.
\newblock \href {https://doi.org/10.18653/v1/W18-6450} {Results of the {WMT}18
  metrics shared task: Both characters and embeddings achieve good
  performance}.
\newblock In \emph{Proceedings of the Third Conference on Machine Translation:
  Shared Task Papers}, pages 671--688, Belgium, Brussels. Association for
  Computational Linguistics.

\bibitem[{Ma et~al.(2019)Ma, Wei, Bojar, and Graham}]{ma-etal-2019-results}
Qingsong Ma, Johnny Wei, Ond{\v{r}}ej Bojar, and Yvette Graham. 2019.
\newblock \href {https://doi.org/10.18653/v1/W19-5302} {Results of the {WMT}19
  metrics shared task: Segment-level and strong {MT} systems pose big
  challenges}.
\newblock In \emph{Proceedings of the Fourth Conference on Machine Translation
  (Volume 2: Shared Task Papers, Day 1)}, pages 62--90, Florence, Italy.
  Association for Computational Linguistics.

\bibitem[{Nakatani et~al.(2008)Nakatani, Amano, Irino, Ishizuka, and
  Kondo}]{NAKATANI2008203}
Tomohiro Nakatani, Shigeaki Amano, Toshio Irino, Kentaro Ishizuka, and Tadahisa
  Kondo. 2008.
\newblock \href {https://doi.org/https://doi.org/10.1016/j.specom.2007.09.003}
  {A method for fundamental frequency estimation and voicing decision:
  Application to infant utterances recorded in real acoustical environments}.
\newblock \emph{Speech Communication}, 50(3):203--214.

\bibitem[{{NLLB Team} et~al.(2022){NLLB Team}, Costa-jussà, Cross, Çelebi,
  Elbayad, Heafield, Heffernan, Kalbassi, Lam, Licht, Maillard, Sun, Wang,
  Wenzek, Youngblood, Akula, Barrault, Mejia-Gonzalez, Hansanti, Hoffman,
  Jarrett, Sadagopan, Rowe, Spruit, Tran, Andrews, Ayan, Bhosale, Edunov, Fan,
  Gao, Goswami, Guzmán, Koehn, Mourachko, Ropers, Saleem, Schwenk, and
  Wang}]{costa2022no}
{NLLB Team}, Marta~R. Costa-jussà, James Cross, Onur Çelebi, Maha Elbayad,
  Kenneth Heafield, Kevin Heffernan, Elahe Kalbassi, Janice Lam, Daniel Licht,
  Jean Maillard, Anna Sun, Skyler Wang, Guillaume Wenzek, Al~Youngblood, Bapi
  Akula, Loic Barrault, Gabriel Mejia-Gonzalez, Prangthip Hansanti, John
  Hoffman, Semarley Jarrett, Kaushik~Ram Sadagopan, Dirk Rowe, Shannon Spruit,
  Chau Tran, Pierre Andrews, Necip~Fazil Ayan, Shruti Bhosale, Sergey Edunov,
  Angela Fan, Cynthia Gao, Vedanuj Goswami, Francisco Guzmán, Philipp Koehn,
  Alexandre Mourachko, Christophe Ropers, Safiyyah Saleem, Holger Schwenk, and
  Jeff Wang. 2022.
\newblock No language left behind: Scaling human-centered machine translation.
\newblock \emph{arXiv preprint arXiv:2207.04672}.

\bibitem[{Ott et~al.(2019)Ott, Edunov, Baevski, Fan, Gross, Ng, Grangier, and
  Auli}]{ott2019fairseq}
Myle Ott, Sergey Edunov, Alexei Baevski, Angela Fan, Sam Gross, Nathan Ng,
  David Grangier, and Michael Auli. 2019.
\newblock fairseq: A fast, extensible toolkit for sequence modeling.
\newblock In \emph{Proceedings of NAACL-HLT 2019: Demonstrations}.

\bibitem[{Panayotov et~al.(2015)Panayotov, Chen, Povey, and
  Khudanpur}]{panayotov2015librispeech}
Vassil Panayotov, Guoguo Chen, Daniel Povey, and Sanjeev Khudanpur. 2015.
\newblock Librispeech: an asr corpus based on public domain audio books.
\newblock In \emph{2015 IEEE international conference on acoustics, speech and
  signal processing (ICASSP)}, pages 5206--5210. IEEE.

\bibitem[{Papineni et~al.(2002)Papineni, Roukos, Ward, and
  Zhu}]{papineni-etal-2002-bleu}
Kishore Papineni, Salim Roukos, Todd Ward, and Wei-Jing Zhu. 2002.
\newblock \href {https://doi.org/10.3115/1073083.1073135} {{B}leu: a method for
  automatic evaluation of machine translation}.
\newblock In \emph{Proceedings of the 40th Annual Meeting of the Association
  for Computational Linguistics}, pages 311--318, Philadelphia, Pennsylvania,
  USA. Association for Computational Linguistics.

\bibitem[{Park and Mulc(2019)}]{park2019css10}
Kyubyong Park and Thomas Mulc. 2019.
\newblock Css10: A collection of single speaker speech datasets for 10
  languages.
\newblock \emph{Interspeech}.

\bibitem[{Polyak et~al.(2021)Polyak, Adi, Copet, Kharitonov, Lakhotia, Hsu,
  Mohamed, and Dupoux}]{polyak21_interspeech}
Adam Polyak, Yossi Adi, Jade Copet, Eugene Kharitonov, Kushal Lakhotia,
  Wei-Ning Hsu, Abdelrahman Mohamed, and Emmanuel Dupoux. 2021.
\newblock {Speech Resynthesis from Discrete Disentangled Self-Supervised
  Representations}.
\newblock In \emph{Proc. Interspeech 2021}.

\bibitem[{Popovi{\'c}(2015)}]{popovic-2015-chrf}
Maja Popovi{\'c}. 2015.
\newblock \href {https://doi.org/10.18653/v1/W15-3049} {chr{F}: character
  n-gram {F}-score for automatic {MT} evaluation}.
\newblock In \emph{Proceedings of the Tenth Workshop on Statistical Machine
  Translation}, pages 392--395, Lisbon, Portugal. Association for Computational
  Linguistics.

\bibitem[{Popovi{\'c}(2017)}]{popovic-2017-chrf}
Maja Popovi{\'c}. 2017.
\newblock \href {https://doi.org/10.18653/v1/W17-4770} {chr{F}++: words helping
  character n-grams}.
\newblock In \emph{Proceedings of the Second Conference on Machine
  Translation}, pages 612--618, Copenhagen, Denmark. Association for
  Computational Linguistics.

\bibitem[{Post(2018)}]{post-2018-call}
Matt Post. 2018.
\newblock \href {https://doi.org/10.18653/v1/W18-6319} {A call for clarity in
  reporting {BLEU} scores}.
\newblock In \emph{Proceedings of the Third Conference on Machine Translation:
  Research Papers}, pages 186--191, Brussels, Belgium. Association for
  Computational Linguistics.

\bibitem[{Pu et~al.(2021)Pu, Chung, Parikh, Gehrmann, and
  Sellam}]{pu-etal-2021-learning}
Amy Pu, Hyung~Won Chung, Ankur Parikh, Sebastian Gehrmann, and Thibault Sellam.
  2021.
\newblock \href {https://doi.org/10.18653/v1/2021.emnlp-main.58} {Learning
  compact metrics for {MT}}.
\newblock In \emph{Proceedings of the 2021 Conference on Empirical Methods in
  Natural Language Processing}, pages 751--762, Online and Punta Cana,
  Dominican Republic. Association for Computational Linguistics.

\bibitem[{Rei et~al.(2021)Rei, Farinha, Zerva, van Stigt, Stewart, Ramos,
  Glushkova, Martins, and Lavie}]{rei-etal-2021-references}
Ricardo Rei, Ana~C Farinha, Chrysoula Zerva, Daan van Stigt, Craig Stewart,
  Pedro Ramos, Taisiya Glushkova, Andr{\'e} F.~T. Martins, and Alon Lavie.
  2021.
\newblock \href {https://aclanthology.org/2021.wmt-1.111} {Are references
  really needed? unbabel-{IST} 2021 submission for the metrics shared task}.
\newblock In \emph{Proceedings of the Sixth Conference on Machine Translation},
  pages 1030--1040, Online. Association for Computational Linguistics.

\bibitem[{Rei et~al.(2020)Rei, Stewart, Farinha, and
  Lavie}]{rei-etal-2020-comet}
Ricardo Rei, Craig Stewart, Ana~C Farinha, and Alon Lavie. 2020.
\newblock \href {https://doi.org/10.18653/v1/2020.emnlp-main.213} {{COMET}: A
  neural framework for {MT} evaluation}.
\newblock In \emph{Proceedings of the 2020 Conference on Empirical Methods in
  Natural Language Processing (EMNLP)}, pages 2685--2702, Online. Association
  for Computational Linguistics.

\bibitem[{Reimers and Gurevych(2019)}]{reimers-2019-sentence-bert}
Nils Reimers and Iryna Gurevych. 2019.
\newblock \href {http://arxiv.org/abs/1908.10084} {Sentence-bert: Sentence
  embeddings using siamese bert-networks}.
\newblock In \emph{Proceedings of the 2019 Conference on Empirical Methods in
  Natural Language Processing}. Association for Computational Linguistics.

\bibitem[{Salesky et~al.(2021{\natexlab{a}})Salesky, M{\"a}der, and
  Klinger}]{salesky2021assessing}
Elizabeth Salesky, Julian M{\"a}der, and Severin Klinger. 2021{\natexlab{a}}.
\newblock Assessing evaluation metrics for speech-to-speech translation.
\newblock In \emph{2021 IEEE Automatic Speech Recognition and Understanding
  Workshop (ASRU)}, pages 733--740. IEEE.

\bibitem[{Salesky et~al.(2021{\natexlab{b}})Salesky, Wiesner, Bremerman,
  Cattoni, Negri, Turchi, Oard, and Post}]{salesky2021mtedx}
Elizabeth Salesky, Matthew Wiesner, Jacob Bremerman, Roldano Cattoni, Matteo
  Negri, Marco Turchi, Douglas~W. Oard, and Matt Post. 2021{\natexlab{b}}.
\newblock Multilingual tedx corpus for speech recognition and translation.
\newblock In \emph{Proceedings of Interspeech}.

\bibitem[{Schneider et~al.(2019)Schneider, Baevski, Collobert, and
  Auli}]{schneider2019wav2vec}
Steffen Schneider, Alexei Baevski, Ronan Collobert, and Michael Auli. 2019.
\newblock wav2vec: Unsupervised pre-training for speech recognition.
\newblock \emph{arXiv preprint arXiv:1904.05862}.

\bibitem[{Schwenk and Douze(2017)}]{schwenk-douze-2017-learning}
Holger Schwenk and Matthijs Douze. 2017.
\newblock \href {https://doi.org/10.18653/v1/W17-2619} {Learning joint
  multilingual sentence representations with neural machine translation}.
\newblock In \emph{Proceedings of the 2nd Workshop on Representation Learning
  for {NLP}}, pages 157--167, Vancouver, Canada. Association for Computational
  Linguistics.

\bibitem[{Sellam et~al.(2020)Sellam, Das, and Parikh}]{sellam-etal-2020-bleurt}
Thibault Sellam, Dipanjan Das, and Ankur Parikh. 2020.
\newblock \href {https://doi.org/10.18653/v1/2020.acl-main.704} {{BLEURT}:
  Learning robust metrics for text generation}.
\newblock In \emph{Proceedings of the 58th Annual Meeting of the Association
  for Computational Linguistics}, pages 7881--7892, Online. Association for
  Computational Linguistics.

\bibitem[{Shimanaka et~al.(2018)Shimanaka, Kajiwara, and
  Komachi}]{shimanaka-etal-2018-ruse}
Hiroki Shimanaka, Tomoyuki Kajiwara, and Mamoru Komachi. 2018.
\newblock \href {https://doi.org/10.18653/v1/W18-6456} {{RUSE}: Regressor using
  sentence embeddings for automatic machine translation evaluation}.
\newblock In \emph{Proceedings of the Third Conference on Machine Translation:
  Shared Task Papers}, pages 751--758, Belgium, Brussels. Association for
  Computational Linguistics.

\bibitem[{Snover et~al.(2006)Snover, Dorr, Schwartz, Micciulla, and
  Makhoul}]{snover-etal-2006-study}
Matthew Snover, Bonnie Dorr, Rich Schwartz, Linnea Micciulla, and John Makhoul.
  2006.
\newblock \href {https://aclanthology.org/2006.amta-papers.25} {A study of
  translation edit rate with targeted human annotation}.
\newblock In \emph{Proceedings of the 7th Conference of the Association for
  Machine Translation in the Americas: Technical Papers}, pages 223--231,
  Cambridge, Massachusetts, USA. Association for Machine Translation in the
  Americas.

\bibitem[{Stanojevi{\'c} et~al.(2015)Stanojevi{\'c}, Kamran, Koehn, and
  Bojar}]{stanojevic-etal-2015-results}
Milo{\v{s}} Stanojevi{\'c}, Amir Kamran, Philipp Koehn, and Ond{\v{r}}ej Bojar.
  2015.
\newblock \href {https://doi.org/10.18653/v1/W15-3031} {Results of the {WMT}15
  metrics shared task}.
\newblock In \emph{Proceedings of the Tenth Workshop on Statistical Machine
  Translation}, pages 256--273, Lisbon, Portugal. Association for Computational
  Linguistics.

\bibitem[{Wang et~al.(2021{\natexlab{a}})Wang, Hsu, Adi, Polyak, Lee, Chen, Gu,
  and Pino}]{wang-etal-2021-fairseq}
Changhan Wang, Wei-Ning Hsu, Yossi Adi, Adam Polyak, Ann Lee, Peng-Jen Chen,
  Jiatao Gu, and Juan Pino. 2021{\natexlab{a}}.
\newblock \href {https://doi.org/10.18653/v1/2021.emnlp-demo.17} {fairseq
  s{\^{}}2: A scalable and integrable speech synthesis toolkit}.
\newblock In \emph{Proceedings of the 2021 Conference on Empirical Methods in
  Natural Language Processing: System Demonstrations}, pages 143--152, Online
  and Punta Cana, Dominican Republic. Association for Computational
  Linguistics.

\bibitem[{Wang et~al.(2022)Wang, Inaguma, Chen, Kulikov, Tang, Hsu, Auli, and
  Pino}]{wang2022simple}
Changhan Wang, Hirofumi Inaguma, Peng-Jen Chen, Ilia Kulikov, Yun Tang,
  Wei-Ning Hsu, Michael Auli, and Juan Pino. 2022.
\newblock Simple and effective unsupervised speech translation.
\newblock \emph{arXiv preprint arXiv:2210.10191}.

\bibitem[{Wang et~al.(2021{\natexlab{b}})Wang, Rivi{\`{e}}re, Lee, Wu,
  Talnikar, Haziza, Williamson, Pino, and
  Dupoux}]{DBLP:conf/acl/WangRLWTHWPD20}
Changhan Wang, Morgane Rivi{\`{e}}re, Ann Lee, Anne Wu, Chaitanya Talnikar,
  Daniel Haziza, Mary Williamson, Juan~Miguel Pino, and Emmanuel Dupoux.
  2021{\natexlab{b}}.
\newblock Voxpopuli: {A} large-scale multilingual speech corpus for
  representation learning, semi-supervised learning and interpretation.
\newblock In \emph{Proceedings of the 59th Annual Meeting of the Association
  for Computational Linguistics and the 11th International Joint Conference on
  Natural Language Processing, {ACL/IJCNLP} 2021, (Volume 1: Long Papers),
  Virtual Event, August 1-6, 2021}, pages 993--1003. Association for
  Computational Linguistics.

\bibitem[{Wang et~al.(2021{\natexlab{c}})Wang, Wu, Gu, and
  Pino}]{wang2021covost}
Changhan Wang, Anne Wu, Jiatao Gu, and Juan Pino. 2021{\natexlab{c}}.
\newblock Covost 2 and massively multilingual speech translation.
\newblock In \emph{Interspeech}, pages 2247--2251.

\bibitem[{Yang et~al.(2020)Yang, Cer, Ahmad, Guo, Law, Constant,
  Hernandez~Abrego, Yuan, Tar, Sung, Strope, and
  Kurzweil}]{yang-etal-2020-multilingual}
Yinfei Yang, Daniel Cer, Amin Ahmad, Mandy Guo, Jax Law, Noah Constant, Gustavo
  Hernandez~Abrego, Steve Yuan, Chris Tar, Yun-hsuan Sung, Brian Strope, and
  Ray Kurzweil. 2020.
\newblock \href {https://doi.org/10.18653/v1/2020.acl-demos.12} {Multilingual
  universal sentence encoder for semantic retrieval}.
\newblock In \emph{Proceedings of the 58th Annual Meeting of the Association
  for Computational Linguistics: System Demonstrations}, pages 87--94, Online.
  Association for Computational Linguistics.

\bibitem[{Zhang* et~al.(2020)Zhang*, Kishore*, Wu*, Weinberger, and
  Artzi}]{Zhang*2020BERTScore:}
Tianyi Zhang*, Varsha Kishore*, Felix Wu*, Kilian~Q. Weinberger, and Yoav
  Artzi. 2020.
\newblock \href {https://openreview.net/forum?id=SkeHuCVFDr} {Bertscore:
  Evaluating text generation with bert}.
\newblock In \emph{International Conference on Learning Representations}.

\end{thebibliography}
\bibliographystyle{acl_natbib}
\appendix

\section{Cross-Modal Evaluation Analysis}
\label{sec:appendix-analysis}

We additionally evaluate \blasersup on different modality combinations when training on speech data only. See the results in Table~\ref{tab:s2st_vs_s2t}. We find that training on speech data only still allows \blaser to obtain similar performance when replacing the reference \audios with text.

\begin{table*}[h!]
\small
    \centering
\sisetup{detect-weight,
         mode=text, 
         table-format = 1.4
         }
    \begin{tabular}{cccSSSSSSSS}
    \toprule
  \multicolumn{3}{l}{\textbf{Modalities}}
  & \textbf{es$\rightarrow$en} & \textbf{ru$\rightarrow$en} & \textbf{hk$\rightarrow$en} & \textbf{fr$\rightarrow$en} & \textbf{en$\rightarrow$de} & \textbf{en$\rightarrow$es} & \textbf{en$\rightarrow$ru}  & \textbf{average}\\ \midrule  
       (Speech,& Speech,& Speech) & \textbf{0.5774} & 0.5347 & 0.6059 & \textbf{0.5730} & 0.3297 &\textbf{0.7512} & \textbf{0.7146} &\textbf{0.5838}\\
       (Speech, &Speech,& Text)   & 0.5588 & \bf 0.5403 &	\bf 0.6093 &	0.5587 &	\bf 0.3426 &	0.7500 &	0.6978 &	0.5796 \\ \bottomrule
    \end{tabular}
    \caption{Pearson's correlation on the test set. Best results are in bold. $(x, y, z)$ indicates the modality used for source input ($x$), translation output ($y$), and reference ($z$). We train \blasersup on speech data only and evaluate the model with references either in speech or text modalities.}
    \label{tab:s2st_vs_s2t}
\end{table*}

\section{Human Evaluation}\label{appendix-sec:human-eval}
\begin{table*}[h!]
    \centering\small
    \begin{tabular}{|l|p{0.8\textwidth}|}\hline
    \rule{0pt}{3ex} Task Descriptions  &
       \begin{itemize}[topsep=0pt,itemsep=1pt]
           \item You will be provided with a pair of audio snippets.
           \item The pair will be in two different languages.
           \item Your task is to assess: (1) if audio1 is coherent; (2) if audio2 is coherent; and (3) how well the pair of audios correspond to each other on a scale from 1-5.
           \item When rating semantic similarity, please ignore minor typos, grammatical errors, and pronunciation errors if they do not affect your understanding of the audio segments.
       \end{itemize} \\ \hline
        \rule{0pt}{3ex} Rating Instructions & \begin{enumerate}[topsep=0pt,itemsep=1pt]
            \item The two sentences are not equivalent, do not share any details, but may be related as pertaining to similar or even different topics.
            \item The two sentences are not equivalent, but share some details. However, some important information differs/is missing, which alters the intent/meaning.
            \item The two sentences are mostly equivalent, but some unimportant details differ. 
            \item The two sentences are equivalent paraphrases of each other. They mean the same with no major or minor differences in meaning, despite potential differences in expression.
            \item The two sentences are exactly and completely equivalent in meaning and usage expression (e.g., formality level, style, multiword expression)
        \end{enumerate}\\\hline
    \end{tabular}
    \caption{Instructions for human evaluations.}
    \label{tab:human_eval_instructions}
\end{table*}

We provide instructions for human evaluations in Table~\ref{tab:human_eval_instructions}.


\section{Ethical Considerations and Limitations}

Translation quality scores were provided by bilingual raters as mentioned in Section \ref{sec:experiments}. They were all paid a fair rate. We can not open-source the data form our experiments given that our sources are shared under \textit{no-derivative} license.

We are evaluating S2ST in an artificial setting given that we have to synthesize the text references. In fact, since there was no metric capable of evaluating the quality in speech, there  was no motivation to build such benchmarks either (the chicken-and-egg problem). However, we expect that next benchmarks for the task will have speech references because of the rise of end-to-end S2ST systems and their quality increase. \blaser{} paves the way so that we can take advantage of such benchmarks when they appear.

Our metric works at the sentence-level, by embedding the entire sentence into an intermediate space. We ignore how sensitive \blaser{} is to the length of the sentence, which is a key aspect when we want to extend to the corpus-level metric in the future. Moreover, we are aware that sometimes sentence embedding do not discriminate different numbers or words that belong to the same word family, which may disregard impactful errors such as the change of a number in the translation output.

\end{document}